\documentclass{article}
\usepackage{iclr2027_conference,times}

\usepackage{amsmath,amsfonts,bm}

\def\eqref#1{equation~\ref{#1}}

\def\1{\bm{1}}

\DeclareMathAlphabet{\mathsfit}{\encodingdefault}{\sfdefault}{m}{sl}
\SetMathAlphabet{\mathsfit}{bold}{\encodingdefault}{\sfdefault}{bx}{n}

\usepackage[hidelinks]{hyperref}
\usepackage{url}
\usepackage{graphicx}
\usepackage{booktabs}
\usepackage{caption}
\usepackage{wrapfig}
\usepackage{multirow}
\usepackage{xcolor}
\usepackage{enumitem}
\usepackage{amsmath,amssymb,amsthm}

\newtheorem{theorem}{Theorem}
\newtheorem*{theoremrestate}{Theorem~\ref{thm:preference_direction}}
\newtheorem*{rewardtheoremrestate}{Theorem~\ref{thm:reward_alignment}}

\newcommand{\cf}{\emph{cf. }}

\newcommand{\pttt}{P-TTT}
\title{\raggedright Using Context Is Not Enough:
Test-Time Training for Personalized Reward Modeling}
\iclrfinalcopy

\author{
\textbf{Bohao Wang}$^{1}$ \quad
\textbf{Xiaoyan Zhao}$^{2}$ \quad
\textbf{Yang Zhang}$^{2}$\thanks{Corresponding authors.} \quad
\textbf{Jinghang Guo}$^{2}$ \\
\textbf{Chun Chen}$^{1}$ \quad
\textbf{Can Wang}$^{1}$ \quad
\textbf{Jiawei Chen}$^{1,*}$ \\
$^{1}$Zhejiang University \\
$^{2}$National University of Singapore
}

\begin{document}
\maketitle
\lhead{Under review as a conference paper at ICLR 2027}

\begin{abstract}
Reinforcement learning from human feedback (RLHF) aligns large language models (LLMs) with human preferences, yet most pipelines learn a single reward model that overlooks individual differences in preferences. Personalized reward models (PRMs) address this by conditioning rewards on user-specific feedback, most commonly through in-context learning (ICL), where a user's historical comparisons are supplied as contextual preference pairs. However, we identify a key limitation of ICL-based PRMs: they fail to capture the preference relations conveyed by contextual pairs. To address this, we propose \textbf{Preference-Aligned Test-Time Training (P-TTT)}, which explicitly encodes these relations into user-specific fast weights for personalized reward prediction. P-TTT introduces sequence-level update and apply operations to match the response-level granularity of preference feedback, together with a preference-aligned objective that directly uses pairwise preference relations to guide fast-weight adaptation. Notably, P-TTT is simple to implement and computationally efficient, updating fast weights within a single forward pass without inference-time backpropagation. Extensive experiments show that P-TTT more effectively captures historical preference relations and outperforms state-of-the-art methods by a large margin.
\end{abstract}

\section{Introduction}
\label{sec:intro}
Reinforcement learning from human feedback (RLHF) plays a central role in aligning large language models (LLMs) with human preferences~\citep{christiano2017deep,ouyang2022training}. Most existing RLHF pipelines learn one shared reward model from human preference annotations over pairs of preferred and rejected responses, and use it to guide policy optimization. Such a formulation implicitly assumes that all humans share a universal preference~\citep{bradley1952rank}. However, this assumption rarely holds in practice: human preferences are inherently heterogeneous, and different users may prefer different responses to the same prompt~\citep{guan2025survey}. A single reward model fitted to such data averages over conflicting preferences, which introduces a systematic bias toward the majority. More importantly, it leaves individual preferences unmodeled, and yet personalization has become essential for modern LLM services~\citep{zhang2024personalization}.

Personalized reward models (PRMs) meet this need by conditioning reward estimation on user-specific information, so that the same response can receive different rewards for different users~\citep{poddar2024personalizing}. The central question is how a PRM leverages user information. Since new users and new feedback arrive continually, a PRM must infer a personalized reward function on the fly from a few historical pairs rather than through per-user retraining. In-context learning (ICL) offers the most straightforward way to do so and has been widely adopted~\citep{zollo2025personalllm,ryan2025synthesizeme}: a user's historical comparisons are concatenated into the prompt as \emph{contextual preference pairs}, from which the reward model is expected to infer the user's preference and adjust its reward for a new \emph{target response} accordingly.

\begin{figure}[t]
    \centering
    \includegraphics[width=\linewidth]{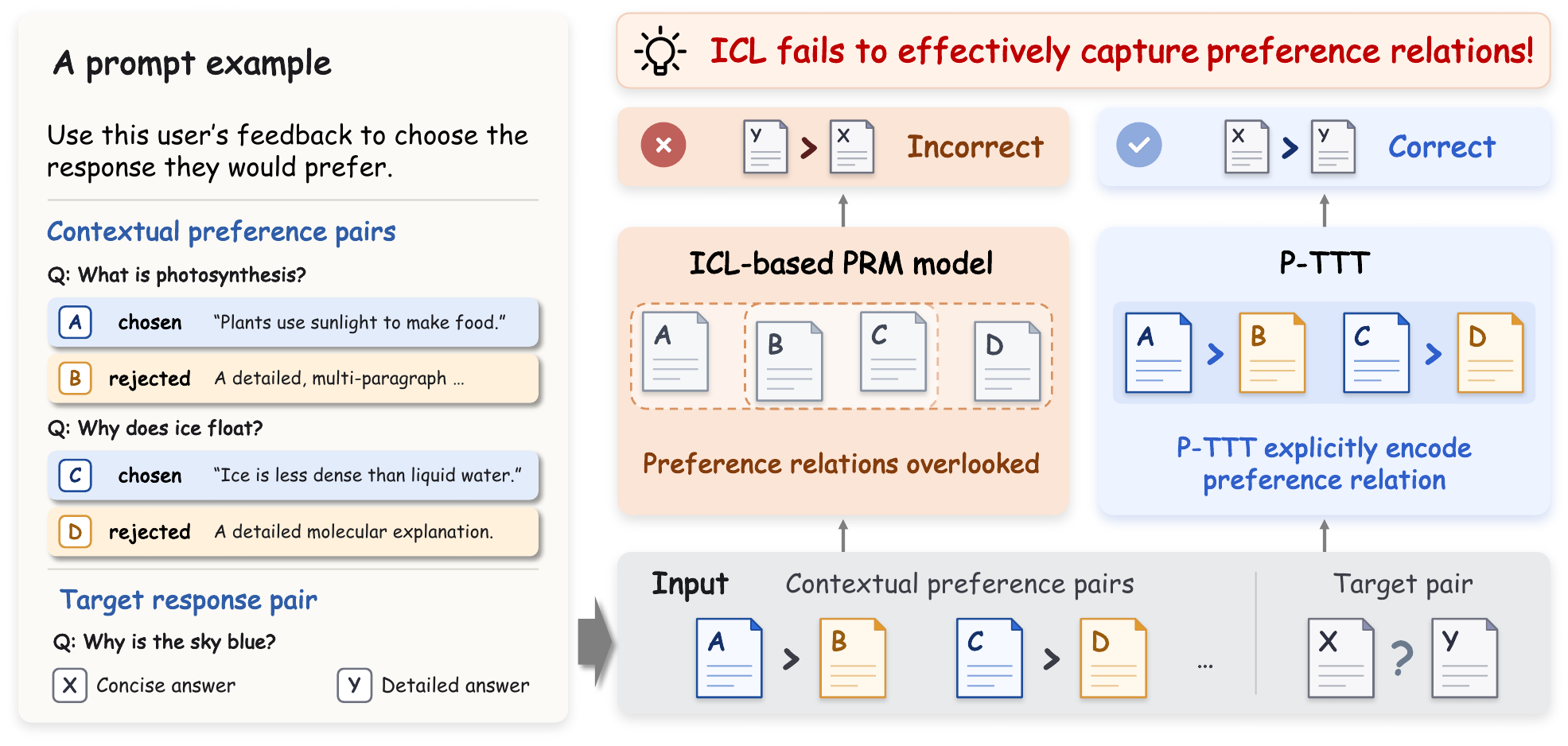}
    \captionsetup{skip=2pt}
    \caption{Illustration of personalized reward modeling with ICL and P-TTT (Ours). While an ICL-based PRM can utilize the content of contextual preference pairs, it may fail to capture their preference relations, resulting in an incorrect target prediction. P-TTT explicitly encodes these relations to better guide target response reward evaluation.}
    \label{fig:intro_overview}
\end{figure}

However, we identify a key limitation of ICL-based PRMs: \textit{they may fail to effectively capture the preference relations conveyed by contextual pairs}. Concretely, they do not reliably distinguish preferred responses from rejected ones (\cf Figure~\ref{fig:intro_overview}). We verify this with a counterfactual test: we reverse the preference direction of the contextual pairs by swapping its chosen and rejected labels, and find that the model's predictions on the same target pairs remain unchanged in 85.95\% of cases on average. The result is striking--- reversed labels describe a user with exactly the opposite interests, yet the model makes almost no corresponding adjustment. It strongly indicates that ICL-based PRMs cannot effectively digest the preference signals expressed in the context. These findings motivate our core research question: \textbf{\textit{How can we enable personalized reward models to better capture preference relations from contextual preference pairs?}}


Towards this end, we propose \textbf{Preference-Aligned Test-Time Training (P-TTT)}. Our approach builds on test-time training~\citep{sun2024learning}, which adapts a small set of \textit{fast weights} at inference time under an explicit learning objective. Unlike ICL, it writes contextual information into parameters rather than inferring it implicitly from the input. Directly applying existing TTT methods to PRMs, however, mismatches the task in two ways: token-level operations are misaligned with response-level preference feedback and reward prediction, while objectives designed for context memorization or language modeling do not explicitly capture preference relations. P-TTT resolves both with two designs: (1) \emph{sequence-level update and apply operations} to align test-time adaptation with the response-level nature of preference annotations and reward prediction; (2) \emph{preference-aligned objective} that directly uses pairwise preference relations as the learning signal for fast-weight adaptation.
Together, these designs enable P-TTT to explicitly internalize such relations and use them to guide target reward prediction.

Notably, P-TTT is easy to implement and computationally efficient.
It reuses the MLP down-projection of a standard reward model as fast weights~\citep{feng2026place}, and updates only this small set of user-specific parameters within a single forward pass, without inference-time backpropagation. 
Moreover, because user history is stored in these parameters, P-TTT avoids including historical examples in the context when scoring target responses, reducing self-attention overhead for lengthy contexts and enabling faster inference than ICL. 
Our theoretical analysis shows that P-TTT captures the direction of preference relations and aligns with the reward modeling objective.
Extensive experiments demonstrate that P-TTT consistently outperforms state-of-the-art baselines by up to 5.85 percentage points in accuracy.

In summary, this work makes the following contributions:
\begin{itemize}[leftmargin=*, labelindent=0pt]
    \item We highlight the importance of modeling preference relations and identify a key limitation of conventional ICL-based strategies: they may fail to effectively capture these relations.
    \item We propose \textbf{Preference-Aligned Test-Time Training (P-TTT)}, a novel strategy for personalized reward modeling that enhances the model’s ability to capture preference relations.
    \item Extensive experiments demonstrate that P-TTT effectively captures preference relations and achieves state-of-the-art performance.
\end{itemize}

\section{Preliminaries}
\label{sec:prelim}

\paragraph{Conventional reward model.}
Reward models are a key component of RLHF for LLMs, providing scalable approximations of human judgments to guide policy optimization while reducing the need for costly manual evaluation.
A reward model $r_\phi(x,y)$ assigns a scalar score to a
response $y$ given a prompt $x$. It is typically implemented as an LLM backbone equipped with a linear reward head~\citep{ouyang2022training}:
\begin{equation}
r_\phi(x,y)
=
\mathbf{v}_r^\top \mathbf{e}_\theta(x,y),
\label{eq:reward_head}
\end{equation}
where $\mathbf{e}_\theta(x,y)$ is the embedding produced by the LLM,
$\mathbf{v}_r$ is the reward-head weight vector, and
$\phi$ denotes the trainable model parameters.

To learn this reward function from pairwise preferences, we consider a dataset
$\mathcal{D}=\{(x_i,y_i^+,y_i^-)\}_{i=1}^{N}$,
where $y_i^+$ and $y_i^-$ are the chosen and rejected responses
to prompt $x_i$, respectively.
The BTL model is then trained by minimizing the negative log-likelihood of the observed preferences:
\begin{equation}
\mathcal{L}_{\mathrm{RM}}(\phi)
=
-\mathbb{E}_{(x,y^+,y^-)\sim\mathcal{D}}
\left[
\log \sigma\left(
r_\phi(x,y^+)-r_\phi(x,y^-)
\right)
\right]
\label{eq:rm_loss}
\end{equation}
where $\sigma(\cdot)$ denotes the sigmoid function. This formulation learns a shared reward function across users,
without explicitly accounting for their individual preferences.

\paragraph{Personalized reward model.}
A personalized reward model extends this formulation by
conditioning the reward function on user-specific information $z$~\citep{poddar2024personalizing}:
\begin{equation}
\mathcal{L}_{\mathrm{PRM}}(\phi) = -\mathbb{E}_{(x,y^+,y^-,z)\sim\mathcal{D}} \left[ \log\sigma\left( r_\phi(x,y^+\mid z)-r_\phi(x,y^-\mid z) \right) \right]. 
\label{eq:personalized}
\end{equation}

In practice, explicit descriptions of user preferences are often
unavailable. Instead, preferences can be inferred
from each user $u$'s historical preference dataset
$\mathcal{D}_u=\{(x_{u,i},y_{u,i}^+,y_{u,i}^-)\}_{i=1}^{m}$,
 and
$y_{u,i}^+$ and $y_{u,i}^-$ are the responses chosen and rejected
by user $u$ for prompt $x_{u,i}$, respectively.
A widely adopted approach to incorporating this information is in-context learning (ICL), where historical examples from $\mathcal{D}_u$ are provided as \emph{contextual preference pairs} to guide reward prediction for a new \emph{target response}~\citep{jincontext,ryan2025synthesizeme,zollo2025personalllm}.
The model is expected to learn to capture the preference information encoded in these contextual pairs through training.


Beyond ICL, other strategies have also been explored, but each faces practical limitations.
\emph{Embedding-based methods}~\citep{poddar2024personalizing} encode users' preference information into compact representations for reward prediction. However, compressing complex preferences into a single embedding may discard fine-grained information. Moreover, LLMs pretrained primarily on natural language may struggle to effectively utilize such representations~\citep{nam2026learning}. Consequently, these methods often exhibit limited performance (\cf Section~\ref{sec:main_results}).
\emph{Parameter-based methods}~\citep{kim2026rethinking,liu2025shared} adapt model parameters to individual users based on their preference feedback. Such methods typically require computationally expensive backpropagation to optimize model parameters, incurring substantial computational overhead. This makes repeated adaptation costly in dynamic settings where user preferences and interactions evolve frequently. Moreover, adapting model parameters from limited user-specific feedback can increase the risk of overfitting.
Further discussion is provided in Appendix~\ref{app}.

\paragraph{Test-Time Training.}
Test-time training (TTT) enables models to adapt to context at inference time by updating a set of parameters $W$, termed \textit{fast weights}~\citep{sun2024learning}.
These weights serve as a memory that retains contextual information for use in subsequent predictions.

Given a sequence of token representations $\{\mathbf{h}_t\}_{t=1}^{T}$, where $\mathbf{h}_t\in\mathbb{R}^{d}$, TTT constructs a query, key, and value $(\mathbf{q}_t,\mathbf{k}_t,\mathbf{v}_t)$ for each token and uses them in two core operations: \textit{update} and \textit{apply}.
Starting from an initialization $W_0$, an \textit{update operation} writes contextual information into the fast weights by learning an association between each key $\mathbf{k}_t$ and its corresponding value $\mathbf{v}_t$:
\begin{equation}
W_t
=
W_{t-1}
-
\eta\left.
\nabla_W \mathcal{L}_{\mathrm{TTT}}
\big(f_W(\mathbf{k}_t),\mathbf{v}_t\big)
\right|_{W=W_{t-1}},
\label{eq:generic_ttt}
\end{equation}
where $\eta$ is the learning rate and
$\mathcal{L}_{\mathrm{TTT}}$ is the adaptation objective.
An \textit{apply operation} then uses the query $\mathbf{q}_t$ to retrieve contextual information stored in the updated fast weights $W_t$ by computing $\mathbf{o}_t=f_{W_t}(\mathbf{q}_t)$.
These operations allow contextual information
to influence subsequent predictions through parameter updates.
However, existing TTT approaches are commonly designed either to memorize contextual information~\citep{sun2024learning} or to predict next tokens~\citep{feng2026place}, rather than modeling pairwise preference relations in PRMs.

\begin{figure}[t]
    \centering
    \includegraphics[width=0.5\linewidth]{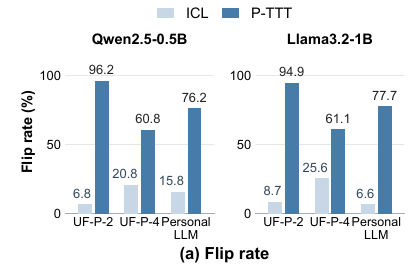}\hfill
    \includegraphics[width=0.5\linewidth]{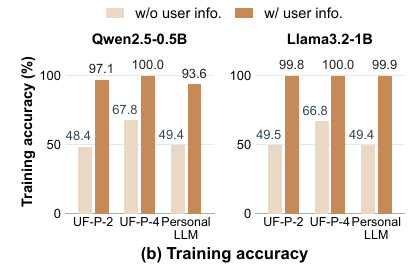}
    \caption{
      (a) Flip rates of ICL and \pttt{}. (b) Training accuracy of ICL-based PRMs with and without contextual preference examples.}
    \label{fig:empirical_exp}
\end{figure}

\section{Empirical Analysis}
\label{sec:motivation}

In this section, we identify a key limitation of ICL-based PRMs: they fail to effectively capture the preference relations expressed in contextual examples. 

\paragraph{Preference-flipping analysis.}
To isolate the role of contextual preference relations, we conduct a counterfactual experiment: we reverse the chosen and rejected labels within selected contextual pairs while leaving the response content unchanged. We then assess whether the model's predictions change in response to these reversals. Specifically, we define the \emph{flip rate} as the fraction of initially correct predictions that change after label reversal. A higher value indicates that contextual preference relations have a greater influence on the model's predictions. Further details are provided in Appendix~\ref{app}. As shown in Figure~\ref{fig:empirical_exp}(a), ICL-based PRMs exhibit an average flip rate of only 14.05\%, retaining their original predictions in most cases. This low flip rate suggests that the relations expressed in the context exert only limited influence on model predictions.

\paragraph{Do the models simply ignore contextual examples?}
The low flip rate may arise if the models make limited use of contextual examples. To examine this possibility, we compare training accuracy with and without contextual preference examples. As shown in Figure~\ref{fig:empirical_exp}(b), the inclusion of these examples yields substantial accuracy gains across all three datasets and both backbones (e.g., on UF-P-2, from 48.44\% to 97.06\% for Qwen and from 49.55\% to 99.81\% for Llama). These results indicate that the models exploit information provided by contextual examples, suggesting that the low flip rate is not attributable solely to a failure to use context.

\paragraph{Conclusion.}
These results reveal a notable discrepancy: ICL-based PRMs can use contextual examples for prediction, yet struggle to capture the preference relations expressed in them.
This suggests that simply placing preference pairs in the context as model input may be insufficient for models to reliably identify the their relations. In particular, the critical preference labels (\emph{chosen} or \emph{rejected}) may not be effectively associated with the corresponding responses, especially when they are embedded in lengthy contextual examples. As a result, rather than modeling the preference relations themselves, models may rely on more readily accessible shortcut cues in historical prompts or responses, such as previously asked questions, to make predictions. Such reliance may limit model generalization to new target samples. 
We also explored simple prompting strategies to mitigate this issue (see the appendix~\ref{app:Prompt_Strategy}), but observed no improvement, suggesting that the difficulty may reflect an intrinsic limitation of how ICL captures preference relations.
These findings motivate us to develop a method that strengthens the preference modeling capability of PRMs.

\section{Method}
\label{sec:method}
\begin{figure}[t]
    \centering
    \includegraphics[width=\linewidth]{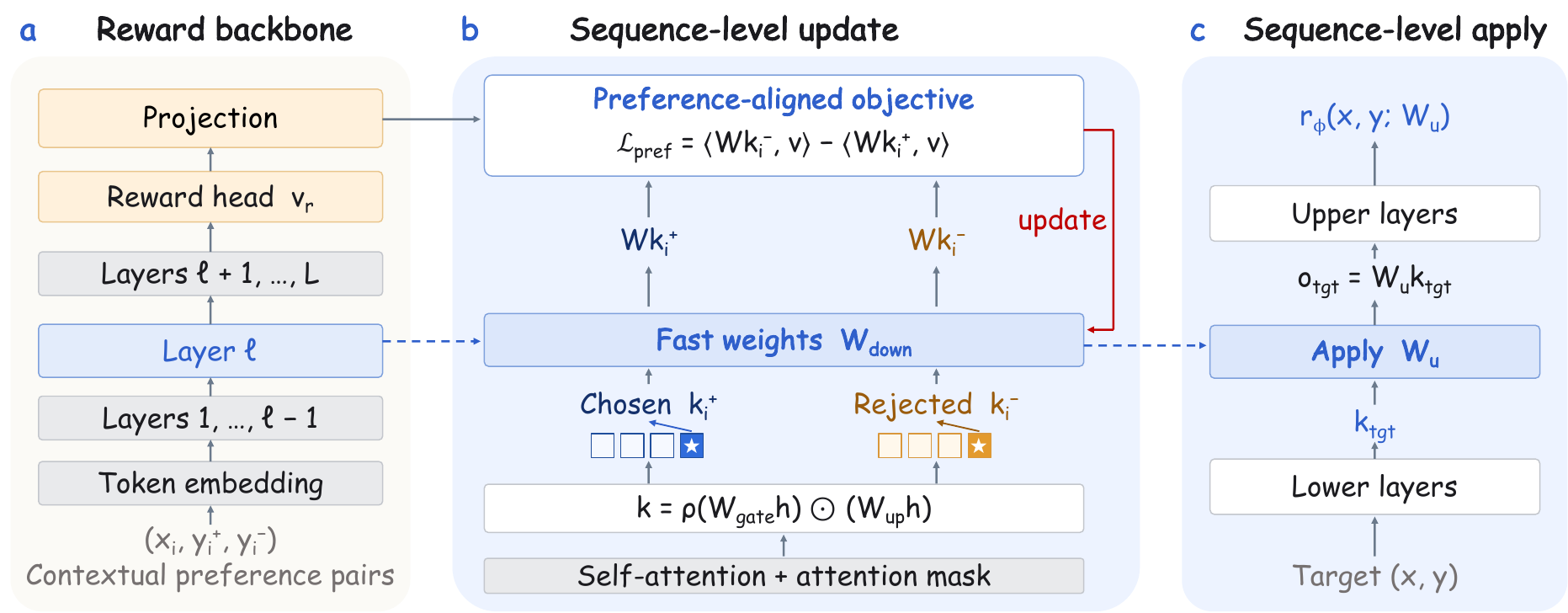}
    \caption{Overview of \pttt{}.}
    \label{fig:method_overview}
\end{figure}

In this section, we propose Preference-Aligned Test-Time Training (\pttt{}) to address the limited ability of ICL to capture preference relations. We first provide an overview of the method (Section~\ref{sec:method_overview}), then describe its two key components: sequence-level update and apply operations (Section~\ref{sec:method_operations}) and a preference-aligned objective (Section~\ref{sec:method_objective}). Finally, we present a theoretical analysis of P-TTT (Section~\ref{sec:method_theory}). Figure~\ref{fig:method_overview} illustrates the overall framework of \pttt{}.

\subsection{Motivation}
\label{sec:method_overview}

\paragraph{Motivation for Test-Time Training.}
Our empirical analysis shows that ICL-based PRMs largely underutilize the preference relations conveyed by contextual preference pairs.
A key limitation is that preference labels are supplied solely as contextual input, requiring the model to infer how they should inform reward predictions for target responses.
This motivates a mechanism that explicitly incorporates contextual preference relations into the model used for prediction.
Test-time training (TTT) offers a natural framework for this purpose by adapting fast weights to contextual examples at inference time.
With an adaptation objective defined over pairwise preferences, these relations can directly guide fast-weight updates.
Building on this insight, we propose \textbf{Preference-Aligned TTT (P-TTT)}, which learns user-specific fast weights from contextual preference pairs for personalized reward prediction.

\paragraph{Adapting TTT to personalized reward modeling.}
However, directly applying TTT to PRMs introduces two key mismatches.
First, a \emph{granularity mismatch}: existing TTT methods typically perform update and apply operations at the token level, whereas preference anontations and reward predictions concern complete responses. We address this mismatch through \emph{sequence-level update and apply operations} (\cf Section \ref{sec:method_operations}), performing both operations on the level of complete response sequences.
Second, an \emph{objective mismatch}: existing TTT methods often use reconstruction objectives to memorize context or LM-aligned objectives for next-token prediction, neither of which explicitly captures preference relations. 
We introduce a \emph{preference-aligned objective} (\cf Section \ref{sec:method_objective}) that makes preference relations an explicit learning signal for fast-weight adaptation. We detail these two designs in the following sections.

\subsection{Sequence-Level Update and Apply Operations}
\label{sec:method_operations}

To resolve the granularity mismatch discussed above, P-TTT both updates and applies the fast weights at the response-sequence level.

\paragraph{User-specific fast weights.}
We repurpose the existing MLP blocks in selected Transformer layers for test-time adaptation without modifying the backbone architecture, following prior work~\citep{feng2026place}. Specifically, given an MLP input $\mathbf{h}\in\mathbb{R}^{d}$, a gated MLP computes
\begin{equation}
    \mathbf{k}
    =
    \rho(W_{\mathrm{gate}}\mathbf{h})
    \odot
    (W_{\mathrm{up}}\mathbf{h}),
    \qquad
    \mathbf{o}
    =
    W_{\mathrm{down}}\mathbf{k},
\end{equation}
where $\rho(\cdot)$ denotes the activation function,
$\mathbf{k}\in\mathbb{R}^{d_{\mathrm{ff}}}$ is the intermediate activation, and $\mathbf{o}\in\mathbb{R}^{d}$ is the MLP output. We use the down-projection matrix $W_{\mathrm{down}}\in\mathbb{R}^{d\times d_{\mathrm{ff}}}$ as the fast weight.
For each user $u$, we initialize the fast weights from the shared model parameters $W_0$ and adapt them using contextual preference pairs $H_u=\{(x_i,y_i^+,y_i^-)\}_{i=1}^{m}$ to obtain the user-specific fast weights $W_u$.

\paragraph{Update operation.}
To align fast-weight adaptation with the response-level nature of preference supervision, we construct each update from complete-response representations rather than token-level.
For each contextual preference pair $(x_i, y_i^+, y_i^-)$, we take the MLP intermediate activations at the final tokens of the chosen and rejected responses as their corresponding keys, denoted by $\mathbf{k}_i^+$ and $\mathbf{k}_i^-$, respectively. Under causal attention, each key summarizes the shared prompt together with its corresponding response, providing a sequence-level representation for adaptation. We further apply an attention mask to prevent cross-response attention, so that the chosen and rejected responses are encoded separately.

Given the two response keys, each contextual preference pair induces a single fast-weight update:
\begin{equation}
    W_i
    =
    W_{i-1}
    -
    \eta
    \left.
    \nabla_W
    \mathcal{L}_{\mathrm{pref}}
    \bigl(
        W;
        \mathbf{k}_i^+,
        \mathbf{k}_i^-,
        \mathbf{v}^+,
        \mathbf{v}^-
    \bigr)
    \right|_{W=W_{i-1}},
\end{equation}
where $\eta$ denotes the test-time learning rate, and $\mathbf{v}^+$ and $\mathbf{v}^-$ are preference-related values associated with the chosen and rejected responses, respectively. The update encodes the preference relation into the fast weights by associating $\mathbf{k}_i^+$ with $\mathbf{v}^+$ and $\mathbf{k}_i^-$ with $\mathbf{v}^-$. We define these values and the preference-aligned objective $\mathcal{L}_{\mathrm{pref}}$ in Section~\ref{sec:method_objective}.

With contextual information encoded in the fast weights $W_u$, we remove them from the input during prediction to reduce the computational cost of attention. Empirically, this removal has a negligible effect on prediction performance (\cf Appendix~\ref{app:context_ablation}).

\paragraph{Apply operation.} 
Similarly, we perform the apply operation on complete-response representations. For each target candidate $(x,y)$, we take the MLP intermediate activation at the final token of the response as the target query $\mathbf{k}_{\mathrm{tgt}}$. We then apply the user-specific fast weights $W_u$ at this position: 
\begin{equation} 
    \mathbf{o}_{\mathrm{tgt}} = W_u\mathbf{k}_{\mathrm{tgt}}. 
\end{equation} 
The preference relations encoded in $W_u$ thus directly modulate the final-token representation used for response-level scoring. To maintain sequence-level adaptation, fast-weight application is restricted to the final-token position, while all other token continue to use the shared weights $W_0$.

\subsection{Preference-Aligned Objective}
\label{sec:method_objective}

To encode preference relations in the fast weights through key--value associations, we first construct a preference-related value from the reward-head weight vector $\mathbf{v}_r$.
The reward head is learned to assign higher rewards to chosen responses than to rejected ones. Its weight vector thus captures a direction associated with higher rewards, making it a natural source of preference-related information.
Accordingly, we define $\mathbf{v}=W_{\mathrm{value}}\mathbf{v}_r$, where $W_{\mathrm{value}}$ is a linear projection, and associate the chosen response's key with $\mathbf{v}^+=+\mathbf{v}$ and the rejected response's key with $\mathbf{v}^-=-\mathbf{v}$.
Specifically, we define the preference-aligned objective as
\begin{equation}
    \mathcal{L}_{\mathrm{pref}}
    \bigl(
        W;\mathbf{k}_i^+,\mathbf{k}_i^-,
        \mathbf{v}^+,\mathbf{v}^-
    \bigr)
    =
    -\langle W\mathbf{k}_i^+,\mathbf{v}^+\rangle
    -\langle W\mathbf{k}_i^-,\mathbf{v}^-\rangle 
    =
    \langle W\mathbf{k}_i^-,\mathbf{v}\rangle
    -
    \langle W\mathbf{k}_i^+,\mathbf{v}\rangle.
\label{eq:preference_aligned_objective}
\end{equation}
Optimizing this objective encourages the fast-weight output of the chosen response to align more strongly with $\mathbf{v}$ than that of the rejected response, thereby encoding the preference relation between the two responses into the fast weights.
The gradient of this objective with respect to the fast weights can be computed efficiently in closed form, yielding the following gradient descent update:
\begin{equation}
    W_i
    =
    W_{i-1}
    +\eta\,\mathbf{v}(\mathbf{k}_i^+)^\top
    -\eta\,\mathbf{v}(\mathbf{k}_i^-)^\top.
    \label{eq:preference_fast_weight_update}
\end{equation}

\paragraph{Training objective.}
To train the model to use the preference information encoded in $W_u$, we optimize the model parameters $\phi$ on target preference pairs using the loss:
\begin{equation}
\mathcal{L}_{\mathrm{train}}(\phi)
=
-\mathbb{E}_{u,(x,y^+,y^-)}
\left[
\log \sigma\left(
r_\phi(x,y^+;W_u)-r_\phi(x,y^-;W_u)
\right)
\right],
\label{eq:training_objective}
\end{equation}
where $r_\phi(x,y;W_u)$ denotes the reward assigned to response $y$ for prompt $x$ using the user-specific fast weights $W_u$.
The two objectives therefore serve complementary roles:
$\mathcal{L}_{\mathrm{pref}}$ guides the encoding of contextual preferences into $W_u$ through fast-weight updates, while $\mathcal{L}_{\mathrm{train}}$ trains the model to use these adapted weights to predict preferences on target pairs.

\subsection{Theoretical Analysis}
\label{sec:method_theory}
Intuitively, P-TTT encodes preference relations by associating each response’s key with a value representing the corresponding preference direction.
In this section, we provide a theoretical analysis of two questions: how reversing contextual preferences affects model predictions, and whether the objective of P-TTT aligns with the reward modeling objective. Detailed assumptions and proofs are provided in Appendices~\ref{app:preference_direction} and~\ref{app:reward_alignment}.

We first establish a direct connection between contextual preference direction and the adaptation-induced change in the target reward margin.

\begin{theorem}
\label{thm:preference_direction}
Let $W_u$ and $W_u^{\mathrm{flip}}$ denote the fast weights adapted from
the same initialization $W_0$ before and after reversing the preference
labels in every contextual pair, respectively.
For any fixed target response pair $(x,y^A,y^B)$, define the reward margin
\begin{equation}
    M(W)=r_\phi(x,y^A;W)-r_\phi(x,y^B;W).
    \label{eq:target_reward_margin}
\end{equation}
Let $\Delta M(W)=M(W)-M(W_0)$ denote the change in reward margin
relative to the initialization. Then reversing all contextual preference
labels negates this change:
\begin{equation}
    \Delta M(W_u^{\mathrm{flip}})=-\Delta M(W_u).
    \label{eq:preference_margin_reversal}
\end{equation}
\end{theorem}

The proof is given in Appendix~\ref{app:preference_direction}.
Theorem~\ref{thm:preference_direction} shows that reversing contextual preferences induces an equal and opposite change to the target reward margin, demonstrating that P-TTT incorporates the direction of contextual preference relations into its predictions.

We next characterize how the preference information encoded in the fast
weights affects the reward of a target response.

\begin{theorem}
\label{thm:reward_alignment}
For a target response $(x,y)$ with query
$\mathbf{k}_{\mathrm{tgt}}$, define its reward correction as
\[
    \Delta r_\phi(x,y)=r_\phi(x,y;W_u)-r_\phi(x,y;W_0).
\]
Under the specified assumptions, with $\eta>0$ and $\mathbf{v}_r\ne\mathbf{0}$,
\begin{align}
\Delta r_\phi(x,y)
&\ge \eta\|\mathbf{v}_r\|_2^2 c_{\mathrm{match}}>0,
\quad \text{if matched to }\mathbf{k}_{i^*}^{+},
\nonumber\\
\Delta r_\phi(x,y)
&\le -\eta\|\mathbf{v}_r\|_2^2 c_{\mathrm{match}}<0,
\quad \text{if matched to }\mathbf{k}_{i^*}^{-}.
\label{eq:expected_reward_alignment}
\end{align}
Here, $i^\star$ indexes the matching contextual response, and
$c_{\mathrm{match}}>0$ denotes the corresponding similarity lower bound.
\end{theorem}

The proof is provided in Appendix~\ref{app:reward_alignment}.
Theorem~\ref{thm:reward_alignment} shows that retrieving a contextual
response from the chosen side increases the target reward, whereas
retrieving one from the rejected side decreases it.
This behavior is directly aligned with the reward modeling objective,
which assigns higher rewards to chosen responses than to rejected ones.
Consequently, when the target responses match contextual keys on their
corresponding preference sides, the P-TTT update increases the reward
margin between the chosen and rejected target responses.

\section{Experiments}

\subsection{Experimental Setup}
\paragraph{Datasets.}
We evaluate on three benchmarks for personalized reward modeling: \textbf{UF-P-2}, \textbf{UF-P-4}, and \textbf{PersonalLLM}, which have been widely used in prior work~\citep{choi2025copl,nam2026learning,kim2026swap}.
UF-P-2 and UF-P-4 construct user preferences along different response quality dimensions \citep{poddar2024personalizing}, while PersonalLLM simulates diverse users through user-specific combinations of pretrained reward models \citep{zollo2025personalllm}.
Across all three benchmarks, the task is to predict a user's preference between two target responses given their historical preference pairs.
Further dataset details and statistics are provided in Appendix~\ref{app:datasets}.

\paragraph{Baselines.}
We compare against eight baselines grouped into five categories.
\textbf{(1) No personalization.} \underline{BTL}~\citep{ouyang2022training} learns a shared reward function without user-specific information.
\textbf{(2) ICL-based PRMs.} \underline{ICL} and \underline{PLUS}~\citep{nam2026learning} condition reward prediction on historical preference pairs or textual summaries of user preferences.
\textbf{(3) Embedding-based PRMs.} \underline{GPO}~\citep{zhao2024group}, \underline{VPL}~\citep{poddar2024personalizing}, and \underline{SPL}~\citep{kim2026swap} encode historical feedback into user-specific representations for preference prediction.
\textbf{(4) Parameter-based PRMs.} \underline{MRM}~\citep{cai2026one} fine-tunes user-specific parameters from a meta-learned initialization using historical preference feedback.
\textbf{(5) Conventional test-time training.} \underline{In-Place TTT}~\citep{feng2026place} is designed for language modeling using a next-token prediction objective.

\paragraph{Implementation Details.}
Following prior reward modeling studies~\citep{nam2026learning}, we use Qwen2.5-0.5B-Instruct~\citep{hui2024qwen2} and Llama3.2-1B-Instruct~\citep{grattafiori2024llama} as reward model backbones for all methods.
For P-TTT, we perform full-parameter fine-tuning using the AdamW optimizer for 3 epochs, with a learning rate of $5 \times 10^{-6}$ and a batch size of 128. During test-time training, we use a learning rate of $\eta = 0.5$, with fast weights introduced every 6 layers.
To ensure fair comparisons, we utilize the source code provided by the original authors and tune the hyperparameters of all baseline methods according to the guidelines specified in their respective publications.
We evaluate performance using pairwise preference accuracy, defined as the proportion of evaluation pairs for which the model assigns a higher reward to the chosen response than to the rejected response, and report the mean accuracy over runs with three random seeds. 

\subsection{Main Results}
\label{sec:main_results}
\begin{table}[!t]
    \captionsetup{skip=2pt}
    \centering
    \caption{Pairwise preference accuracy (\%). We report the mean and standard deviation over 3 seeds. Best results are bold.}
    \label{tab:main_results}
    \footnotesize
    \definecolor{venuecolor}{HTML}{526D82}
    \setlength{\tabcolsep}{2pt}
    \renewcommand{\arraystretch}{1.12}
    \begin{tabular*}{\linewidth}{@{\extracolsep{\fill}}llccc@{}}
        \toprule
        Model & Method & UF-P-2 & UF-P-4 & PersonalLLM \\
        \midrule
        \multirow{9}{*}{Qwen2.5-0.5B} & BTL\,{\scriptsize\textcolor{venuecolor}{(NIPS 2022)}} & \makebox[2.5em][r]{49.88}\,$\pm$\,\makebox[2.5em][l]{0.12} & \makebox[2.5em][r]{58.35}\,$\pm$\,\makebox[2.5em][l]{0.84} & \makebox[2.5em][r]{48.83}\,$\pm$\,\makebox[2.5em][l]{1.03} \\
         & ICL & \makebox[2.5em][r]{53.33}\,$\pm$\,\makebox[2.5em][l]{1.92} & \makebox[2.5em][r]{60.46}\,$\pm$\,\makebox[2.5em][l]{0.31} & \makebox[2.5em][r]{51.05}\,$\pm$\,\makebox[2.5em][l]{1.05} \\
         & PLUS\,{\scriptsize\textcolor{venuecolor}{(ICLR 2026)}} & \makebox[2.5em][r]{51.76}\,$\pm$\,\makebox[2.5em][l]{0.07} & \makebox[2.5em][r]{58.86}\,$\pm$\,\makebox[2.5em][l]{0.39} & \makebox[2.5em][r]{50.65}\,$\pm$\,\makebox[2.5em][l]{0.48} \\
         & GPO\,{\scriptsize\textcolor{venuecolor}{(ICLR 2024)}} & \makebox[2.5em][r]{50.52}\,$\pm$\,\makebox[2.5em][l]{0.07} & \makebox[2.5em][r]{55.44}\,$\pm$\,\makebox[2.5em][l]{0.64} & \makebox[2.5em][r]{50.25}\,$\pm$\,\makebox[2.5em][l]{0.05} \\
         & VPL\,{\scriptsize\textcolor{venuecolor}{(NIPS 2024)}} & \makebox[2.5em][r]{50.16}\,$\pm$\,\makebox[2.5em][l]{0.07} & \makebox[2.5em][r]{57.73}\,$\pm$\,\makebox[2.5em][l]{1.09} & \makebox[2.5em][r]{50.07}\,$\pm$\,\makebox[2.5em][l]{0.06} \\
         & SPL\,{\scriptsize\textcolor{venuecolor}{(ICLR 2026)}} & \makebox[2.5em][r]{52.68}\,$\pm$\,\makebox[2.5em][l]{1.98} & \makebox[2.5em][r]{58.80}\,$\pm$\,\makebox[2.5em][l]{0.40} & \makebox[2.5em][r]{50.18}\,$\pm$\,\makebox[2.5em][l]{0.03} \\
         & MRM\,{\scriptsize\textcolor{venuecolor}{(SIGIR 2026)}} & \makebox[2.5em][r]{69.55}\,$\pm$\,\makebox[2.5em][l]{0.37} & \makebox[2.5em][r]{59.40}\,$\pm$\,\makebox[2.5em][l]{0.39} & \makebox[2.5em][r]{58.23}\,$\pm$\,\makebox[2.5em][l]{0.29} \\
         & In-Place TTT\,{\scriptsize\textcolor{venuecolor}{(ICLR 2026)}} & \makebox[2.5em][r]{50.32}\,$\pm$\,\makebox[2.5em][l]{0.56} & \makebox[2.5em][r]{52.73}\,$\pm$\,\makebox[2.5em][l]{0.61} & \makebox[2.5em][r]{50.07}\,$\pm$\,\makebox[2.5em][l]{1.15} \\
         & \textbf{P-TTT} & \textbf{\boldmath \makebox[2.5em][r]{75.40}\,$\pm$\,\makebox[2.5em][l]{0.54}} & \textbf{\boldmath \makebox[2.5em][r]{64.00}\,$\pm$\,\makebox[2.5em][l]{0.57}} & \textbf{\boldmath \makebox[2.5em][r]{61.27}\,$\pm$\,\makebox[2.5em][l]{0.55}} \\
        \midrule
        \multirow{9}{*}{Llama3.2-1B} & BTL\,{\scriptsize\textcolor{venuecolor}{(NIPS 2022)}} & \makebox[2.5em][r]{49.92}\,$\pm$\,\makebox[2.5em][l]{0.07} & \makebox[2.5em][r]{58.47}\,$\pm$\,\makebox[2.5em][l]{0.10} & \makebox[2.5em][r]{49.97}\,$\pm$\,\makebox[2.5em][l]{0.06} \\
         & ICL & \makebox[2.5em][r]{51.88}\,$\pm$\,\makebox[2.5em][l]{0.50} & \makebox[2.5em][r]{60.08}\,$\pm$\,\makebox[2.5em][l]{0.82} & \makebox[2.5em][r]{56.02}\,$\pm$\,\makebox[2.5em][l]{3.72} \\
         & PLUS\,{\scriptsize\textcolor{venuecolor}{(ICLR 2026)}} & \makebox[2.5em][r]{51.36}\,$\pm$\,\makebox[2.5em][l]{0.28} & \makebox[2.5em][r]{58.67}\,$\pm$\,\makebox[2.5em][l]{0.37} & \makebox[2.5em][r]{50.80}\,$\pm$\,\makebox[2.5em][l]{1.34} \\
         & GPO\,{\scriptsize\textcolor{venuecolor}{(ICLR 2024)}} & \makebox[2.5em][r]{50.76}\,$\pm$\,\makebox[2.5em][l]{0.42} & \makebox[2.5em][r]{56.57}\,$\pm$\,\makebox[2.5em][l]{0.16} & \makebox[2.5em][r]{50.10}\,$\pm$\,\makebox[2.5em][l]{0.09} \\
         & VPL\,{\scriptsize\textcolor{venuecolor}{(NIPS 2024)}} & \makebox[2.5em][r]{50.24}\,$\pm$\,\makebox[2.5em][l]{0.21} & \makebox[2.5em][r]{58.57}\,$\pm$\,\makebox[2.5em][l]{0.30} & \makebox[2.5em][r]{50.02}\,$\pm$\,\makebox[2.5em][l]{0.03} \\
         & SPL\,{\scriptsize\textcolor{venuecolor}{(ICLR 2026)}} & \makebox[2.5em][r]{50.60}\,$\pm$\,\makebox[2.5em][l]{0.32} & \makebox[2.5em][r]{59.20}\,$\pm$\,\makebox[2.5em][l]{0.33} & \makebox[2.5em][r]{50.13}\,$\pm$\,\makebox[2.5em][l]{0.06} \\
         & MRM\,{\scriptsize\textcolor{venuecolor}{(SIGIR 2026)}} & \makebox[2.5em][r]{70.71}\,$\pm$\,\makebox[2.5em][l]{0.39} & \makebox[2.5em][r]{62.29}\,$\pm$\,\makebox[2.5em][l]{0.68} & \makebox[2.5em][r]{57.40}\,$\pm$\,\makebox[2.5em][l]{0.35} \\
         & In-Place TTT\,{\scriptsize\textcolor{venuecolor}{(ICLR 2026)}} & \makebox[2.5em][r]{50.28}\,$\pm$\,\makebox[2.5em][l]{0.87} & \makebox[2.5em][r]{52.77}\,$\pm$\,\makebox[2.5em][l]{0.97} & \makebox[2.5em][r]{49.52}\,$\pm$\,\makebox[2.5em][l]{0.43} \\
         & \textbf{P-TTT} & \textbf{\boldmath \makebox[2.5em][r]{74.48}\,$\pm$\,\makebox[2.5em][l]{0.18}} & \textbf{\boldmath \makebox[2.5em][r]{64.60}\,$\pm$\,\makebox[2.5em][l]{0.97}} & \textbf{\boldmath \makebox[2.5em][r]{62.50}\,$\pm$\,\makebox[2.5em][l]{0.26}} \\
        \bottomrule
    \end{tabular*}
\end{table}

As shown in Table~\ref{tab:main_results}, P-TTT achieves the highest mean accuracy across all three datasets with both backbones, outperforming the strongest baseline in each setting by an average of 3.94 percentage points.
Among ICL-based methods, ICL’s weaker performance is consistent with our analysis in Section~\ref{sec:motivation}: its difficulty in capturing preference relations limits its ability to infer user interests and generalize effectively. Although PLUS learns a summarizer, compressing feedback into textual summaries may still discard fine-grained preference information. Embedding-based methods also offer limited gains over BTL, potentially due to information loss when encoding complex users' histories into compact representations. MRM benefits from meta-learned adaptation, but constraining adaptation to low-dimensional combinations of shared reward functions may limit its ability to capture complex preferences patterns. Finally, the poor performance of In-Place TTT is consistent with the mismatch between its language modeling aligned objective and the goal of PRMs. In contrast, P-TTT directly aligns adaptation with preference relations, enabling the model to capture useful preference information from user feedback and better generalize to unseen samples.

To further evaluate whether P-TTT captures preference relations, we revisit the preference-flipping experiment in Section~\ref{sec:motivation}. As shown in Figure~\ref{fig:empirical_exp}(a), P-TTT achieves substantially higher flip rates than ICL across all settings, with an average of 77.81\% versus 14.05\% for ICL. These results suggest that P-TTT’s predictions are more responsive to changes in the preference relations expressed in user feedback. These findings provide evidence that P-TTT more effectively models user-specific preference relations.

\subsection{Ablation Studies}

\begin{wraptable}{r}{0.48\textwidth}
    \vspace{-10pt}
    \centering
    \captionsetup{skip=2pt}
    \vspace{0pt}
    \centering
    \caption{Ablation study. SLUA: sequence-level update and apply operations; PAO: preference-aligned objective. Best results are bold.}
    \label{tab:ablation}
    \footnotesize
    \setlength{\tabcolsep}{2pt}
    \renewcommand{\arraystretch}{1.12}
    \begin{tabular*}{\linewidth}{@{\extracolsep{\fill}}lccc@{}}
        \toprule
        Method & UF-P-2 & UF-P-4 & PersonalLLM \\
        \midrule
        BTL & 49.88 & 58.35 & 48.83 \\
        ICL & 53.33 & 60.46 & 51.05 \\
        \midrule
        w/o SLUA
            & 59.50 & 50.24 & 53.65 \\
        w/o PAO
            & 50.12 & 58.19 & 49.95 \\
        \textbf{P-TTT}
            & \textbf{75.40} & \textbf{64.00} & \textbf{61.27} \\
        \bottomrule
    \end{tabular*}
    \vspace{-8pt}
\end{wraptable}

In this section, we ablate the two core components of P-TTT using Qwen2.5-0.5B-Instruct as the backbone: \textit{sequence-level update and apply operations} (SLUA) and the \textit{preference-aligned objective} (PAO).
For \textit{w/o SLUA}, we perform fast-weight updates at every token of the contextual responses and apply the adapted weights at every token of the target responses.
For \textit{w/o PAO}, we replace the preference-aligned update objective with the LM-aligned objective used in In-Place TTT.
As shown in Table~\ref{tab:ablation}, removing either component reduces accuracy across all three benchmarks.
Without SLUA, accuracy drops substantially (e.g. 64.00\% to 50.24\% on UF-P-4), supporting the importance of matching adaptation granularity to response-level preference feedback.
Without PAO, accuracy falls to near-BTL levels across all datasets (e.g. 75.40\% to 50.12\% on UF-P-2), suggesting that updates under a misaligned objective fail to encode useful user-specific preference information into the fast weights.
These results highlight the complementary contributions of P-TTT’s two modules and underscore the importance of aligning both the granularity and the objective of adaptation with user-specific preferences.


\subsection{Efficiency Comparison}
\label{sec:efficiency}

\begin{wrapfigure}{r}{0.48\textwidth}
    \vspace{-10pt}
    \centering
    \includegraphics[width=\linewidth]{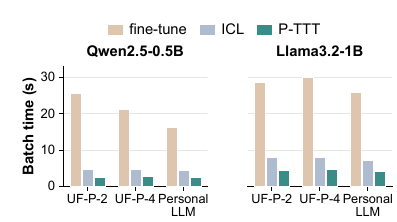}
    \captionsetup{skip=2pt}
    \caption{Inference time per batch samples for different methods. }
    \label{fig:efficiency}
    \vspace{-8pt}
\end{wrapfigure}

In this section, we evaluate the inference efficiency of P-TTT against ICL and per-user fine-tuning (\cf Appendix~\ref{app:finetuning_strategy} for details), as shown in Figure~\ref{fig:efficiency}. We measure the total time required to evaluate rewards for a batch of 32 samples.
P-TTT is consistently the fastest across both backbones and all three datasets, requiring 2.446--4.694\,s per batch and achieving 1.70--1.83$\times$ and 6.21--9.72$\times$ speedups over ICL and fine-tuning, respectively.
Unlike per-user fine-tuning, P-TTT updates fast weights within a single forward pass without inference-time backpropagation.
Moreover, encoding contextual information in fast weights allows P-TTT to omit lengthy contextual preference examples from the model input during target prediction, reducing attention computation relative to ICL.
These results demonstrate that P-TTT achieves strong predictive performance while maintaining high inference efficiency.


\section{Conclusion}
In this paper, we identify a key limitation of ICL-based personalized reward models: they underutilize the preference relations expressed in contextual examples. To address this limitation, we propose Preference-Aligned Test-Time Training (\pttt{}), which explicitly encodes these relations into user-specific fast weights through sequence-level update and apply operations and a preference-aligned objective. 
Empirical results demonstrate that P-TTT outperforms existing methods while reducing inference latency, highlighting its effectiveness and efficiency for personalized reward modeling.
Future work could explore continual personalization, enabling LLMs to adapt to evolving user preferences while selectively retaining and updating long-term user information.

\bibliographystyle{iclr2027_conference}
\bibliography{iclr2027_conference}

\clearpage

\appendix
\section{Related Work}
\label{app:related_work}

\paragraph{Personalized Reward Model}
Personalized reward models (PRMs) capture heterogeneous user preferences rather than learning a single shared reward function~\citep{poddar2024personalizing,shenfeld2025language}. Existing approaches broadly fall into three categories:
\textbf{(1) Embedding-based methods} encode historical feedback into compact user representations for reward prediction \citep{poddar2024personalizing,zhao2024group,kim2026swap}. However, compressing complex preference patterns can obscure fine-grained information, while LLM-based reward models pretrained on natural language may struggle to utilize these embeddings effectively~\citep{nam2026learning}.
\textbf{(2) Parameter-based methods} adapt model parameters using individual feedback~\citep{kim2026rethinking,liu2025shared}. Full-model adaptation incurs substantial inference-time overhead due to per-user backpropagation. Parameter-efficient approaches reduce this cost by representing preferences in a low-dimensional space and optimizing a small set of user-specific parameters~\citep{shenfeld2025language,cai2026one}. However, this restriction may limit expressiveness. Moreover, when user feedback is scarce, these methods are prone to overfitting the observed feedback, limiting generalization to unseen inputs.
\textbf{(3) ICL-based methods} provide historical preference pairs as textual context~\citep{jincontext,ryan2025synthesizeme,zollo2025personalllm}, with related approaches learning compact, interpretable user summaries ~\citep{nam2026learning}. These methods rely on the model to infer how historical comparisons should influence new judgments, while lengthy histories increase processing costs. Prior work reports limited performance of ICL-based methods and broadly attributes this limitation to models’ difficulty in leveraging lengthy contexts~\citep{nam2026learning}. 
Our work focuses on this ICL-based PRM setting and provides a more fine-grained analysis: trained models can leverage contextual content while making limited use of the preference relations between chosen and rejected responses in historical pairs.

\paragraph{Test Time Training}
Test-time training (TTT) enables models to adapt to context through inference-time parameter updates.
TTT layers and neural memory architectures incorporate contextual information into fast weights through self-supervised updates~\citep{sun2024learning,behrouz2026titans,liu2026test}.
Recent work further aligns adaptation with language modeling: TTT-E2E optimizes next-token prediction over the observed context and meta-learns an initialization for adaptation~\citep{tandon2512end}, while In-Place TTT repurposes existing MLP down-projections as fast weights with an LM-aligned objective and efficient chunk-wise updates~\citep{feng2026place}.
TTT-NTP uses next-position contextual hidden states to supervise local fast-weight updates in pretrained LLMs, connecting adaptation more directly to next-token prediction~\citep{ouyang2026test}.
Context-specific gradient updates have also been shown to improve long-context performance by addressing limitations of static self-attention~\citep{bansal2026let}.
Beyond language modeling, test-time alignment has been explored in sequential recommendation to track user interest shifts through incremental self-supervised updates~\citep{zhang2025test}.
These studies motivate adaptation objectives tailored to the target task.
Our work focuses on personalized reward modeling, where adaptation must capture comparative preferences over complete responses~\citep{poddar2024personalizing,nam2026learning}, and uses historical chosen--rejected pairs to guide fast-weight updates.


\section{Proof of Theorem~\ref{thm:preference_direction}}
\label{app:preference_direction}
For analytical tractability, following \citet{feng2026place}, we focus our analysis on a single Transformer block equipped with P-TTT.

\begin{theoremrestate}
Let $W_u$ and $W_u^{\mathrm{flip}}$ denote the fast weights adapted from
the same initialization $W_0$ before and after reversing the preference
labels in every contextual pair, respectively.
For any fixed target response pair $(x,y^A,y^B)$, define the reward margin
\begin{equation}
    M(W)=r_\phi(x,y^A;W)-r_\phi(x,y^B;W).
\end{equation}
Let $\Delta M(W)=M(W)-M(W_0)$ denote the change in reward margin
relative to the initialization. Then reversing all contextual preference
labels negates this change:
\begin{equation}
    \Delta M(W_u^{\mathrm{flip}})=-\Delta M(W_u).
\end{equation}
\end{theoremrestate}

\begin{proof}
Let $\mathbf{k}(x,y)$ denote the MLP intermediate activation at the final
token of response $y$ to prompt $x$.
For contextual pair $i$, let
$\mathbf{k}_i^\pm=\mathbf{k}(x_i,y_i^\pm)$ and define
$\mathbf{d}_i=\mathbf{k}_i^+-\mathbf{k}_i^-$.
The preference-aligned objective satisfies
\[
    \mathcal{L}_{\mathrm{pref}}
    (W;\mathbf{k}_i^+,\mathbf{k}_i^-,
    \mathbf{v}^+,\mathbf{v}^-)
    =-\mathbf{v}^\top W\mathbf{d}_i,
    \qquad
    \nabla_W\mathcal{L}_{\mathrm{pref}}
    =-\mathbf{v}\mathbf{d}_i^\top.
\]
Thus, accumulating the $m$ updates yields
\begin{equation}
    W_u-W_0
    =\sum_{i=1}^{m}(W_i-W_{i-1})
    =\eta\mathbf{v}\sum_{i=1}^{m}\mathbf{d}_i^\top.
    \label{eq:direction_accumulated_update}
\end{equation}
Reversing all contextual preference labels replaces each
$\mathbf{d}_i$ by $-\mathbf{d}_i$.
Consequently,
\[
    W_u^{\mathrm{flip}}-W_0
    =-\eta\mathbf{v}\sum_{i=1}^{m}\mathbf{d}_i^\top
    =-(W_u-W_0).
\]
For the fixed target pair, let
\[
    \mathbf{k}_{\mathrm{tgt}}^A=\mathbf{k}(x,y^A),
    \qquad
    \mathbf{k}_{\mathrm{tgt}}^B=\mathbf{k}(x,y^B),
    \qquad
    \Delta\mathbf{k}_{\mathrm{tgt}}
    =\mathbf{k}_{\mathrm{tgt}}^A-\mathbf{k}_{\mathrm{tgt}}^B.
\]
The reward model gives
\begin{equation}
    r_\phi(x,y;W)-r_\phi(x,y;W_0)
    =\mathbf{v}_r^\top(W-W_0)\mathbf{k}_{\mathrm{tgt}}.
    \label{eq:direction_linear_readout}
\end{equation}
It follows that
\[
    \Delta M(W)
    =\mathbf{v}_r^\top(W-W_0)\Delta\mathbf{k}_{\mathrm{tgt}}.
\]
Since $\Delta\mathbf{k}_{\mathrm{tgt}}$ is unchanged under contextual label reversal,
\begin{align*}
    \Delta M(W_u^{\mathrm{flip}})
    &=\mathbf{v}_r^\top(W_u^{\mathrm{flip}}-W_0)
      \Delta\mathbf{k}_{\mathrm{tgt}}\\
    &=-\mathbf{v}_r^\top(W_u-W_0)
      \Delta\mathbf{k}_{\mathrm{tgt}}\\
    &=-\Delta M(W_u).\qedhere
\end{align*}
\end{proof}

\section{Proof of Theorem~\ref{thm:reward_alignment}}
\label{app:reward_alignment}

\begin{rewardtheoremrestate}
For a target response $(x,y)$ with query
$\mathbf{k}_{\mathrm{tgt}}$, define its reward correction as
\[
    \Delta r_\phi(x,y)=r_\phi(x,y;W_u)-r_\phi(x,y;W_0).
\]
Under the specified assumptions, with $\eta>0$ and $\mathbf{v}_r\ne\mathbf{0}$,
\begin{align}
\Delta r_\phi(x,y)
&\ge \eta\|\mathbf{v}_r\|_2^2 c_{\mathrm{match}}>0,
\quad \text{if matched to }\mathbf{k}_{i^*}^{+},
\nonumber\\
\Delta r_\phi(x,y)
&\le -\eta\|\mathbf{v}_r\|_2^2 c_{\mathrm{match}}<0,
\quad \text{if matched to }\mathbf{k}_{i^*}^{-}.
\end{align}
Here, $i^*$ indexes the matching contextual response, and
$c_{\mathrm{match}}>0$ denotes the corresponding similarity lower bound.
\end{rewardtheoremrestate}

\paragraph{Assumption.}
We assume that $W_{\mathrm{value}}$ is the identity transformation,
so that $\mathbf{v}=\mathbf{v}_r$.
We further make the following two assumptions.

\textit{(i) Positive similarity to a matching key.}
For the key of target response $\mathbf{k}_{\mathrm{tgt}}$, there exist
an index $i^*\in\{1,\ldots,m\}$, a label $s\in\{+,-\}$,
and a constant $c_{\mathrm{match}}>0$ such that
\begin{equation}
    \langle\mathbf{k}_{i^*}^{s},\mathbf{k}_{\mathrm{tgt}}\rangle
    \ge c_{\mathrm{match}}.
    \label{eq:target_key_matching}
\end{equation}

\textit{(ii) Zero aggregate contribution from the remaining keys.}
The remaining entries satisfy
\begin{equation}
    \sum_{\substack{1\le i\le m,\ t\in\{+,-\}\\(i,t)\ne(i^*,s)}}
    \langle\mathbf{k}_i^t,\mathbf{k}_{\mathrm{tgt}}\rangle
    \mathbf{v}^t=\mathbf{0}.
    \label{eq:retrieval_interference}
\end{equation}

\begin{proof}
Equation~\ref{eq:preference_fast_weight_update} implies
\[
    W_u-W_0
    =\sum_{i=1}^{m}(W_i-W_{i-1})
    =\eta\sum_{i=1}^{m}\sum_{t\in\{+,-\}}
    \mathbf{v}^t(\mathbf{k}_i^t)^\top.
\]
Consequently,
\begin{equation}
    \Delta\mathbf{o}_{\mathrm{tgt}}
    :=(W_u-W_0)\mathbf{k}_{\mathrm{tgt}}
    =\eta\sum_{i=1}^{m}\sum_{t\in\{+,-\}}
    \langle\mathbf{k}_i^t,\mathbf{k}_{\mathrm{tgt}}\rangle\mathbf{v}^t.
    \label{eq:preference_value_retrieval}
\end{equation}
By Equation~\ref{eq:retrieval_interference},
\[
    \Delta\mathbf{o}_{\mathrm{tgt}}
    =\eta\mathbf{v}^{s}
    \langle\mathbf{k}_{i^*}^{s},\mathbf{k}_{\mathrm{tgt}}\rangle.
\]
Then
\begin{equation}
    \Delta r_\phi(x,y)
    =\mathbf{v}_r^\top\Delta\mathbf{o}_{\mathrm{tgt}}
    =\eta(\mathbf{v}_r^\top\mathbf{v}^{s})
    \langle\mathbf{k}_{i^*}^{s},\mathbf{k}_{\mathrm{tgt}}\rangle.
    \label{eq:retrieved_value_reward}
\end{equation}
Using $\mathbf{v}^+=\mathbf{v}_r$ and $\mathbf{v}^-=-\mathbf{v}_r$,
Equations~\ref{eq:target_key_matching} and~\ref{eq:retrieved_value_reward} give
\begin{align*}
    \Delta r_\phi(x,y)
    &\ge \eta\|\mathbf{v}_r\|_2^2 c_{\mathrm{match}}>0,
    \qquad s=+,\\
    \Delta r_\phi(x,y)
    &\le -\eta\|\mathbf{v}_r\|_2^2 c_{\mathrm{match}}<0,
    \qquad s=-.\qedhere
\end{align*}
\end{proof}

\section{Preference-Flipping Analysis}
\label{app}
\label{app:flip_protocol}

This section provides experimental details for the preference-flipping analysis in Section~\ref{sec:motivation} and Figure~\ref{fig:empirical_exp}(a).
We conduct a counterfactual experiment to examine whether the model changes its prediction when contextual preference relations are reversed while the response content is kept unchanged. 

\paragraph{Label reversal.}
For an evaluation sample $j$, let $H_j$ denote its contextual preference examples, and let $(x_j,y_j^A,y_j^B)$ denote the target prompt and its two candidate responses.
We construct a counterfactual context $\widetilde H_j^{(v)}$ by swapping the chosen and rejected labels within selected contextual pairs, without necessarily reversing all pairs.
The contextual prompts and response content, as well as the target prompt and both candidate responses, remain unchanged.

To ensure that the contextual label changes imply a reversal of the target preference, we validate each variant using the dataset-defined preference rules.
We first verify that the original history implies an unambiguous target preference.
We then identify all personas or generated users consistent with the modified history.
A variant is retained only if this set is nonempty and every compatible user prefers the opposite target response.
All other variants are excluded.
Let $\mathcal{V}_j$ denote the resulting set of valid label-reversal variants for sample $j$.

\paragraph{Flip rate.}
Let $z_j\in\{0,1\}$ denote the original target preference label, with $z_j=1$ indicating that $y_j^A$ is preferred to $y_j^B$ and $z_j=0$ indicating the reverse.
Let $\hat z_j$ be the model's prediction under $H_j$, and $\hat z_j^{(v)}$ its prediction under $\widetilde H_j^{(v)}$.
For each valid variant, the target preference label is $1-z_j$.
We evaluate samples that the model initially predicts correctly and that have at least one valid label-reversal variant:
\begin{equation}
    \mathcal{C}
    =\{j:\hat z_j=z_j,\ |\mathcal{V}_j|>0\}.
\end{equation}
For each $j\in\mathcal{C}$, we compute its flip rate as the fraction of valid variants for which the prediction changes to the other candidate:
\begin{equation}
    \mathrm{Flip\ Rate}_j
    =\frac{1}{|\mathcal{V}_j|}
    \sum_{v\in\mathcal{V}_j}
    \mathbf{1}\!\left[\hat z_j^{(v)}=1-z_j\right],
\end{equation}
where $\mathbf{1}[\cdot]$ is the indicator function.
Since the original prediction is correct and each valid variant reverses the target preference, this prediction change agrees with the reversed target preference label.
We then average the per-sample flip rates equally:
\begin{equation}
    \mathrm{Flip\ Rate}
    =\frac{1}{|\mathcal{C}|}
    \sum_{j\in\mathcal{C}}\mathrm{Flip\ Rate}_j.
    \label{eq:flip_rate}
\end{equation}
Each initially correct sample with at least one valid variant receives equal weight, regardless of its number of valid variants.
A higher flip rate indicates that contextual preference relations have a greater influence on the model's predictions.

\section{Limitations of Prompt-Based Strategies}
\label{app:Prompt_Strategy}
We investigate whether explicit instructions can improve the model's use of contextual preference relations, evaluating Qwen2.5-0.5B-Instruct on UF-P-2 and UF-P-4. Specifically, we prepend an instruction emphasizing the preference labels to the contextual preference examples during both training and evaluation:
\begin{quote}
\small
Pay explicit attention to the preference labels in every historical example. `Chosen response' is the response this user preferred; `Rejected response' is the response this user liked less. These labels, not the order in which responses appear, specify the user's preference. Compare the chosen and rejected responses to infer which traits the user favors and disfavors. When scoring the response to the current request, reward alignment with the labeled chosen responses and penalize alignment with the labeled rejected responses. Treat the labels shown in the examples as the evidence of this user's preferences.
\end{quote}

As shown in Table~\ref{tab:prompt_strategy}, explicitly emphasizing preference labels does not yield consistent improvements. On UF-P-2, the instruction modestly increases the flip rate from 6.83\% to 8.98\% and test accuracy from 53.33\% to 53.97\%. On UF-P-4, however, the flip rate decreases from 20.82\% to 8.17\%, while test accuracy drops from 60.46\% to 58.80\%. These results suggest that, in the settings evaluated, explicit instructions alone are insufficient to reliably improve the model's use of contextual preference relations.

\begin{table}[htbp]
    \captionsetup{skip=2pt}
    \centering
    \caption{Effect of explicit preference-label instructions on flip rate and test accuracy (\%) using Qwen2.5-0.5B-Instruct.}
    \label{tab:prompt_strategy}
    \small
    \setlength{\tabcolsep}{10pt}
    \begin{tabular}{llcc}
        \toprule
        Method & Metric & UF-P-2 & UF-P-4 \\
        \midrule
        ICL & Flip rate  & 6.83 & 20.82 \\
            & Test accuracy  & 53.33 & 60.46 \\
        \midrule
        ICL + instruction & Flip rate  & 8.98 & 8.17 \\
                          & Test accuracy  & 53.97 & 58.80 \\
        \bottomrule
    \end{tabular}
\end{table}

\section{Details of Per-User Fine-Tuning}
\label{app:finetuning_strategy}

In this section, we describe the experimental setup for per-user fine-tuning.
We fine-tune the reward model using the $m$ historical preference pairs associated with each target sample.
Let $H_u=\{(x_i,y_i^+,y_i^-)\}_{i=1}^{m}$ denote these preference pairs.
The tuning objective is
\begin{equation}
    \mathcal{L}_{\mathrm{FT}}(\phi;H_u)
    =-\frac{1}{m}\sum_{i=1}^{m}
    \log\sigma\!\left(r_\phi(x_i,y_i^+)-r_\phi(x_i,y_i^-)\right).
\end{equation}
The target pair and its preference label are excluded from adaptation.
After fine-tuning, the model scores the target responses without contextual examples in the input.
We reset the model parameters and optimizer state before adapting to each target sample, including samples from the same user.
We evaluate fine-tuning for 3 optimization steps for each sample.

\section{Retaining Context in Self-Attention}
\label{app:context_ablation}

In this section, we examine whether P-TTT benefits from retaining historical context in the input during target prediction, to assess whether the adapted fast weights effectively retain contextual preference information.
The variant \textit{P-TTT w/ context} retains contextual tokens so that target responses can attend to them through self-attention, in addition to using the adapted fast weights.
As shown in Table~\ref{tab:context_ablation}, retaining context has little effect on prediction accuracy.
These results suggest that P-TTT effectively encodes and retains relevant contextual preference information in its fast weights, supporting its design of omitting contextual tokens during target prediction.

\begin{table}[htbp]
    \captionsetup{skip=2pt}
    \centering
    \caption{Accuracy (\%) with context accessible to self-attention.}
    \label{tab:context_ablation}
    \small
    \setlength{\tabcolsep}{10pt}
    \renewcommand{\arraystretch}{1.12}
    \begin{tabular}{lccc}
        \toprule
        Method & UF-P-2 & UF-P-4 & PersonalLLM \\
        \midrule
        P-TTT & 75.40 & 64.00 & 61.27 \\
        P-TTT w/ context & 75.84 & 64.06 & 61.10 \\
        \bottomrule
    \end{tabular}
\end{table}

\section{Dataset Overview and Statistics}
\label{app:datasets}

We evaluate on three personalized preference datasets: UF-P-2, UF-P-4, and PersonalLLM. For each dataset, we reserve 100 preference pairs as a shared survey pool and exclude them from the training targets.
For each user--target sample, we sample contextual pairs from this pool and label them according to the user's preferences. The experiments use $m=4$ contextual pairs.
Table~\ref{tab:dataset_statistics} summarizes their statistics. 

\begin{table}[htbp]
    \captionsetup{skip=2pt}
    \centering
    \caption{Processed dataset statistics.}
    \label{tab:dataset_statistics}
    \small
    \setlength{\tabcolsep}{7pt}
    \begin{tabular}{llrrr}
        \toprule
        Dataset & Split & Users & Samples & $m$ \\
        \midrule
        UF-P-2 & Train & 2 & 7,516 & 4 \\
               & Test & 2 & 832 & 4 \\
        \midrule
        UF-P-4 & Train & 4 & 15,740 & 4 \\
               & Test & 4 & 1,660 & 4 \\
        \midrule
        PersonalLLM & Train & 100 & 10,000 & 4 \\
                    & Test & 120 & 2,000 & 4 \\
        \bottomrule
    \end{tabular}
\end{table}

\paragraph{UF-P-2 and UF-P-4.}
Following UF-P~\citep{poddar2024personalizing}, we construct user preferences from UltraFeedback ratings along different response quality dimensions.
UF-P-2 includes two user types that prefer \emph{helpfulness} and \emph{honesty}, respectively; UF-P-4 adds two types that prefer \emph{instruction following} and \emph{truthfulness}.
Each user prefers the response with the higher rating on their designated dimension.
Training and testing share the same user types but use disjoint target prompts.

\paragraph{PersonalLLM.}
PersonalLLM~\citep{zollo2025personalllm} simulates diverse users through user-specific weighted combinations of ten reward models, with pairwise preferences determined by the resulting scores.
We use 100 users for training and evaluate on both these users and 20 unseen users.
The test set contains 1,000 samples from each group, and we report accuracy over the combined 2,000 samples.

\end{document}